\pdfoutput=1
\PassOptionsToPackage{table}{xcolor}
\documentclass[11pt]{article}
\usepackage{graphicx}
\usepackage{xcolor}   
\usepackage{calc} 
\usepackage{colortbl}  
\usepackage{booktabs}
\usepackage{caption}
\usepackage{tabulary}          
\usepackage{cuted}  
\usepackage{caption}
\usepackage{dblfloatfix}
\usepackage[final]{acl}
\usepackage{makecell}
\usepackage{times}
\usepackage{pifont}
\usepackage{tabularx} 
\usepackage{float}
\usepackage{latexsym}
\usepackage{academicons}
\usepackage{algorithm}
\usepackage{amsmath}
\usepackage[most]{tcolorbox}
\usepackage[font=small]{caption}
\usepackage{algpseudocode}
\usepackage[T1]{fontenc}

\usepackage[utf8]{inputenc}

\usepackage{microtype}
\usepackage{fontawesome}
\usepackage{inconsolata}

\usepackage{multirow}
\newcommand{\cmark}{\textcolor{green}{\ding{51}}}
\newcommand{\xmark}{\textcolor{red}{\ding{55}}}
\newcommand{\methodname}{IDEAlign}

\title{\methodname{}: Comparing Ideas of Large Language Models to Domain Experts}

\author{
Hyunji Nam \and Lucia Langlois \and James Malamut \\
\textbf{Mei Tan \and Dorottya Demszky} \\
Stanford University \\
\texttt{\{hjnam, ddemszky\}@stanford.edu}
}

\begin{document}
\maketitle
\begin{abstract}
Large language models (LLMs) are increasingly used to produce open-ended, interpretive annotations, yet there is no validated, scalable measure of \emph{idea-level similarity} to expert annotations. We (i) introduce the content evaluation of LLM annotations as a core, understudied task, (ii) propose \methodname{} for capturing expert similarity judgments via \emph{pick-the-odd-one-out} tasks, and (iii) benchmark various similarity methods (text embeddings, topic models, and LLM-as-a-judge) against these human ratings. Applying this approach to two real-world educational datasets (e.g., interpreting math reasoning and feedback generation), we find that most metrics fail to capture the nuanced dimensions of similarity meaningful to experts. LLM-as-a-judge performs best (11–18\% improvement over other methods) but still falls short of expert alignment, making it useful as a triage tool rather than a substitute for human review. Our work demonstrates the difficulty of evaluating open-ended LLM annotations at scale, and positions \methodname{} as a reusable protocol for benchmarking on this task to help guide responsible deployment of LLMs.\footnote{\href{https://github.com/EduNLP/IDEAlign}{https://github.com/EduNLP/IDEAlign}}
\end{abstract}

\section{Introduction}
Comparing model predictions to human labels has long been central to model evaluation. When outputs are categorical, assessment is relatively straightforward~\citep{doi:10.1126/science.adi6000, zhou-etal-2024-llm, strachan2024testing, bojic2025comparing, schroeder2025trustllmjudgmentsreliability}. By contrast, large language models (LLMs) are increasingly used to generate unstructured, open-ended annotations involving reasoning and interpretation grounded in domain-specific knowledge~\citep{tan2024largelanguagemodelsdata, yu2023temporaldatameetsllm}. In socially impactful domains, like education, LLM-based tools like KhanMigo and Brisk observe student thinking and writing to provide insights to students and teachers~\citep{10.1145/3637528.3671498, graesser2004autotutor, khan2023harnessing, team2024learnlm, wang2024bridgingnoviceexpertgapmodels}. Because these interpretive model outputs can shape instruction and learning, expert-grounded evaluation of LLMs is essential. However, doing so is non-trivial: the tasks require domain and context knowledge (e.g., standards and learning objectives), and rarely admit a single ``correct" response.

\begin{figure*}[t]
  \centering
  \includegraphics[height=6cm]{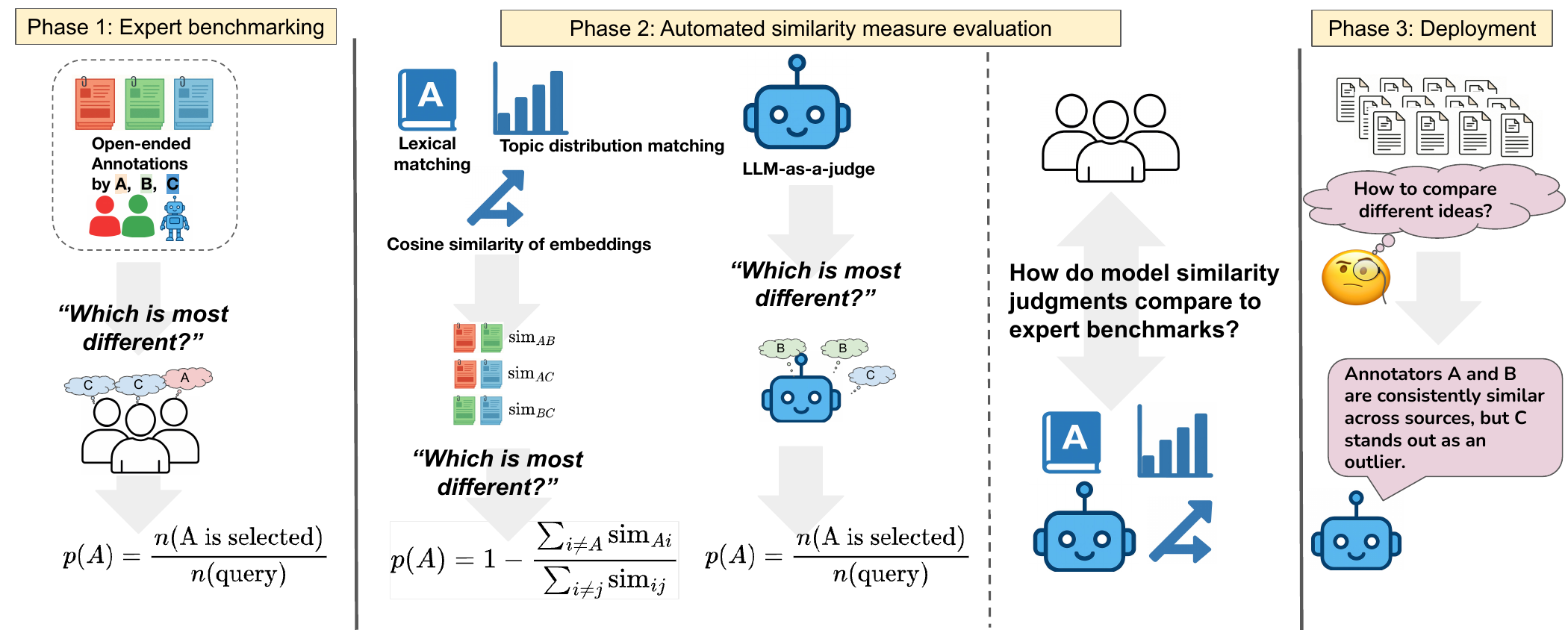}
  \caption{IDEAlign is a framework for evaluating alignment between LLM-generated and expert annotations on open-ended, interpretive tasks. It consists of three stages: 
  \textbf{(1) Benchmarking} expert similarity judgments via \emph{odd-one-out} tasks, 
  \textbf{(2) Validating} automated (model) similarity methods (e.g., lexical \protect\raisebox{-0.2ex}{\includegraphics[height=1em]{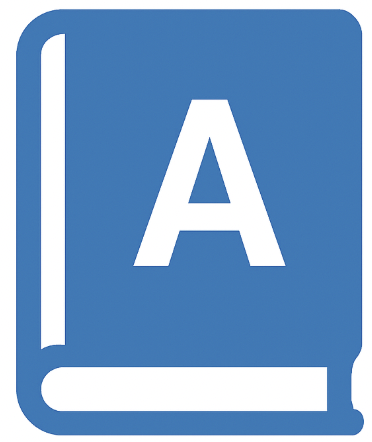}}, 
  embedding-based \protect\raisebox{-0.2ex}{\includegraphics[height=1em]{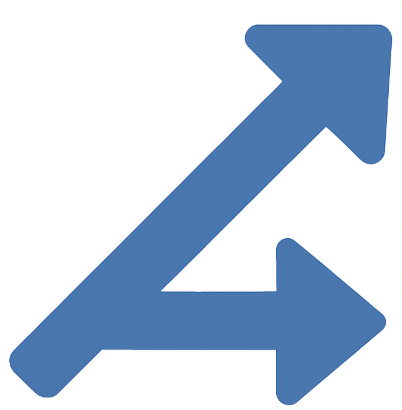}}, 
  topic-based \protect\raisebox{-0.2ex}{\includegraphics[height=1em]{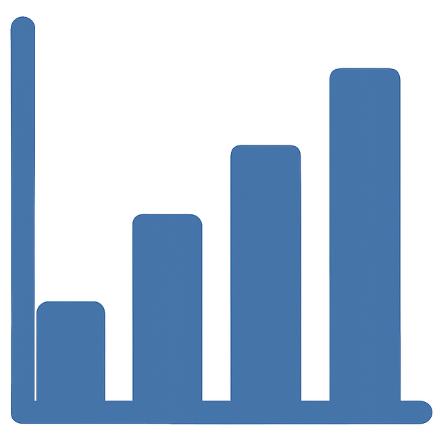}}, 
  and LLM-as-a-judge \protect\raisebox{-0.2ex}{\includegraphics[height=1em]{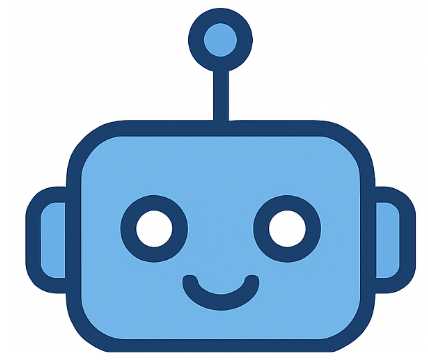}}) 
  against expert benchmarks by comparing answer distributions, and 
  \textbf{(3) Deploying} the best-validated model to assess similarity of ideas generated by LLMs and domain experts at scale.}
  \label{fig:main}
\end{figure*}

To evaluate and improve LLMs on such tasks, we need to assess their alignment with experts in terms of ideas: whether they express the same underlying interpretations or recommended actions, rather than imitating surface features (tone, length, phrasing).  Measuring idea-level similarity is difficult because (i) expert annotations legitimately vary across perspectives and communities~\citep{pmlr-v202-santurkar23a}; and (ii) absolute scales for subjective constructs are difficult to elicit reliably~\citep{Annett01112002}. These challenges motivate a relative judgment approach for eliciting human judgments, which can serve as benchmarks. 

Existing evaluations of interpretive LLM annotations remain limited: many focus on predefined code sets~\citep{wadhwa2024revisitingrelationextractionera, larionov-etal-2019-semantic} or prediction accuracy over reasoning quality~\citep{yu2023temporaldatameetsllm}.
Prior work on general long-form texts have targeted measuring and improving quality of text generation along dimensions of self-consistency~\citep{huang-etal-2023-large} and human values, such as creativity and truthfulness~\citep{ismayilzada2025evaluatingcreativeshortstory, spangher-etal-2024-llms, tan2024reframing, lin2021truthfulqa}, but these approaches are not suited for assessing conceptual alignment between LLM and expert annotations. To our knowledge, alignment of ideas on open-ended, interpretive annotations has not been systematically studied.

Automated text comparison metrics have complementary limitations: lexical overlap (e.g., BLEU~\citep{papineni-etal-2002-bleu} ignores paraphrase; embedding-based~\citep{reimers2019sentencebertsentenceembeddingsusing, Radford2019LanguageMA, openaiada2022, neelakantan2022textcodeembeddingscontrastive, pillutla2021mauvemeasuringgapneural} and topic-based
measures~\citep{grootendorst2022bertopic, NIPS2001_296472c9} conflate content with style/length and offer limited control over the dimension of comparison. LLM-as-a-judge~\citep{zheng2023judgingllmasajudgemtbenchchatbot} can provide a promising alternative but requires careful evaluation, as LLMs show sensitivity to prompt templates~\citep{wei2025systematicevaluationllmasajudgellm} and stylistic biases~\citep{feuer2025style}. However, these methods have yet to be validated against human similarity judgments for open-ended, interpretive tasks.

\textbf{We introduce \methodname{}, a benchmarking framework that elicits expert similarity judgments via \emph{pick-the-odd-one-out} tasks.} Given three annotations of the same source document, experts select the most dissimilar one along a specified, domain-relevant criterion (e.g., ``What change to the student's writing is being recommended?"). This triplet design follows prior comparative judgment and ``topic intruder" work~\citep{blei2009}, leveraging the well-established reliability of relative judgments, which dates back from Thurstone's \emph{Law of Comparative Judgment}~\citep{thurstone-comparative} and the influential Bradley-Terry model~\citep{bradley1952rank} to more recent research and applied studies involving subjective evaluation~\citep{laming2003human, blei2009, tarricone2016using, routh2023rating}, such as the widely adopted Adaptive Comparative Judgment~\citep{pollitt2012method} in educational settings, where teachers or peer evaluators compare the relative quality of two pieces of work rather than judging each one individually. ~\citet{pollitt2012method} makes an argument that relying on implicit judgments of quality through pairwise comparisons with ``intrinsically simple" task can give reliable results without having to specify individual criteria or calibrate subjective scores across many evaluators. 
In \methodname{}, expert selections yield a probability distribution over the “odd” item; model predictions are compared to this distribution using Hellinger distance.

Using \methodname{}, we evaluate families of automated similarity measures: (i) lexical overlap (BLEU), (ii) cosine similarity from 16 embedding models with/without PCA-based post-processing, (iii) topic-distribution divergence, where the topics are found using topic modeling methods like BERTopic and MALLET-LDA, and (iv) LLM-as-a-judge. Methods other than LLM-as-a-judge can only output pairwise similarity scores rather than directly answering the odd-one-out task, so we convert the models' pairwise scores to the probability of each item being selected in a triplet (Figure~\ref{fig:main}).

We study two education tasks where expert, idea-level interpretation matters: (1) assessing students' mathematical reasoning from classroom transcripts, and (2) providing inline feedback to student essays. Across both settings, most similarity measures fail to produce answers that agree with experts, and in particular, they show sensitivity to confounders like style and length. By contrast, \textbf{prompting LLMs to answer the odd-one-out task directly yields the strongest agreement with experts (11–18\% improvement over other methods)}. Since LLM-as-a-judge still falls short of expert agreement, we position it as a helpful tool that can reduce human workload rather than a substitute for expert review. 

Our work demonstrates the difficulty of evaluating open-ended LLM annotations at scale, and
positions \methodname{} as a reusable protocol for comparing model and human similarity judgments. Our goal is to inform responsible deployment of LLMs in education and beyond.

\section{Related Work}
\paragraph{Comparing LLMs to humans in open-ended annotations.} LLMs are moving beyond tasks like storytelling and dialogue generation~\citep{ismayilzada2025evaluatingcreativeshortstory, wiegreffe-etal-2022-reframing, ouyang2022traininglanguagemodelsfollow} toward more complex, interpretive work that requires data synthesis and reasoning based on domain knowledge~\citep{tan2024largelanguagemodelsdata, yu2023temporaldatameetsllm,spangher-etal-2024-llms}. For example, in educational settings, LLM-based tools are used to assess student thinking and generate feedback as teachers~\citep{10.1145/3637528.3671498, graesser2004autotutor, khan2023harnessing, team2024learnlm, wang2024bridgingnoviceexpertgapmodels}. However, most evaluation of LLM-generated annotations is designed for classification tasks with pre-defined coding schemes~\citep{wadhwa2024revisitingrelationextractionera, larionov-etal-2019-semantic} or for assessing tone~\citep{tan2024reframing} or writing quality along dimensions, like helpfulness and creativity~\citep{ismayilzada2025evaluatingcreativeshortstory, spangher-etal-2024-llms, wiegreffe-etal-2022-reframing}. However, for open-ended annotations, there is no single correct answer to compare against, and the quality of ideas is challenging to evaluate with ratings. Other works on evaluation focus on lexical features, length, and grammar~\citep{Mu_oz_Ortiz_2024, martinez2025beware}, but ignore content alignment, which is critical for assessing the reliability of LLM-generated annotations for tasks that require domain expertise. As an alternative to assuming a single ground truth, MAUVE~\citep{pillutla2021mauvemeasuringgapneural} compares the distributions of human and LLM-generated texts using KL divergence. However, their method relies on text embeddings to estimate these distributions and is therefore constrained by the performance of embedding models, which are sensitive to length and style. Crucially, to our knowledge, the alignment of ideas between LLMs and experts on open-ended, interpretive annotations has not been investigated.

\paragraph{Automated similarity metrics.} Many standard NLP metrics have been developed for text comparison such as BLEU~\cite{papineni-etal-2002-bleu} and ROUGE~\cite{lin-2004-rouge}. However, their focus on lexical overlap fails to capture similarity in underlying ideas. \textbf{Cosine similarity of text embeddings} is also a popular method, and different embedding models have been developed and studied, including Word2Vec~\cite{ng-abrecht-2015-better}, BERT~\cite{devlin-etal-2019-bert, zhao-etal-2019-moverscore, zhang2020bertscoreevaluatingtextgeneration}, Sentence-BERT~\cite{reimers2019sentencebertsentenceembeddingsusing}, GPT-2~\cite{Radford2019LanguageMA}, GTE~\cite{li2023generaltextembeddingsmultistage}, and proprietary models like ada~\cite{openaiada2022} and cpt text~\cite{neelakantan2022textcodeembeddingscontrastive}. ~\citet{mu2018allbutthetopsimpleeffectivepostprocessing} proposed post-processing based on PCA (removing projections of the top principal components from word embeddings) to improve performance on NLP semantic similarity benchmarks. However, it remains unclear how well these embedding-based similarity scores align with expert judgments. In fact, ~\citet{fabbri-etal-2021-summeval} shows that many models show poor correlations with human evaluation. ~\citet{muennighoff-etal-2023-mteb} also shows that no single embedding model consistently achieves the best performance across NLP tasks.

An alternative to embeddings is to represent annotations as \textbf{topic distributions} using methods like LDA~\cite{NIPS2001_296472c9} and BERTopic~\cite{grootendorst2022bertopic}. However, as our experiments will show, depending on the number of topics, the representations may be too coarse to yield meaningful comparisons, especially when most annotations in our datasets discuss the same broad topic.

\section{\methodname{}}

We propose \methodname{}, a benchmarking method for eliciting expert similarity judgments and evaluating automated measures against human benchmarks. 
\subsection{Eliciting expert similarity judgments via \emph{odd-one-out} tasks}
A naive way of collecting expert data is to ask them for continuous similarity ratings on a fixed scale for every pair of annotations. However, absolute judgments are cognitively difficult and may yield noisy, hard-to-calibrate labels. Prior work has therefore relied on relative judgments as an alternative to absolute scoring~\citep{laming2003human, jones2012summative,  pollitt2012method, tarricone2016using, routh2023rating}. Our task design builds on this line of work, including the ``topic intruder/odd-one-out" paradigm introduced by ~\citet{blei2009}. 

In the \emph{odd-one-out} task, evaluators are provided a group of three annotations $A, B, C$ written about the same source text. They select which annotation is most different along domain-specific similarity criteria established by experts. We repeat this evaluation step with many triplets and multiple evaluators to aggregate answers. For each triplet $(A, B, C)$, this yields an empirical distribution that represents how likely each item $x \in \{A, B, C\}$ is to be selected as most dissimilar:
\begin{equation}\label{eq:human_chosen}
P(x|A,B,C) = \frac{\#(x \text{ is picked})}{\#(A, B, C \text{ occurs in triplet})}
\end{equation} This allows us to construct an evaluation dataset that accounts for human uncertainty in identifying an outlier. We compare model predictions to this distribution using Hellinger distance. We additionally ask the evaluators to rate the difficulty of picking the odd one out from each triplet on a scale of easy, moderate, and difficult.

\begin{figure}[htbp]
  \includegraphics[width=\columnwidth]{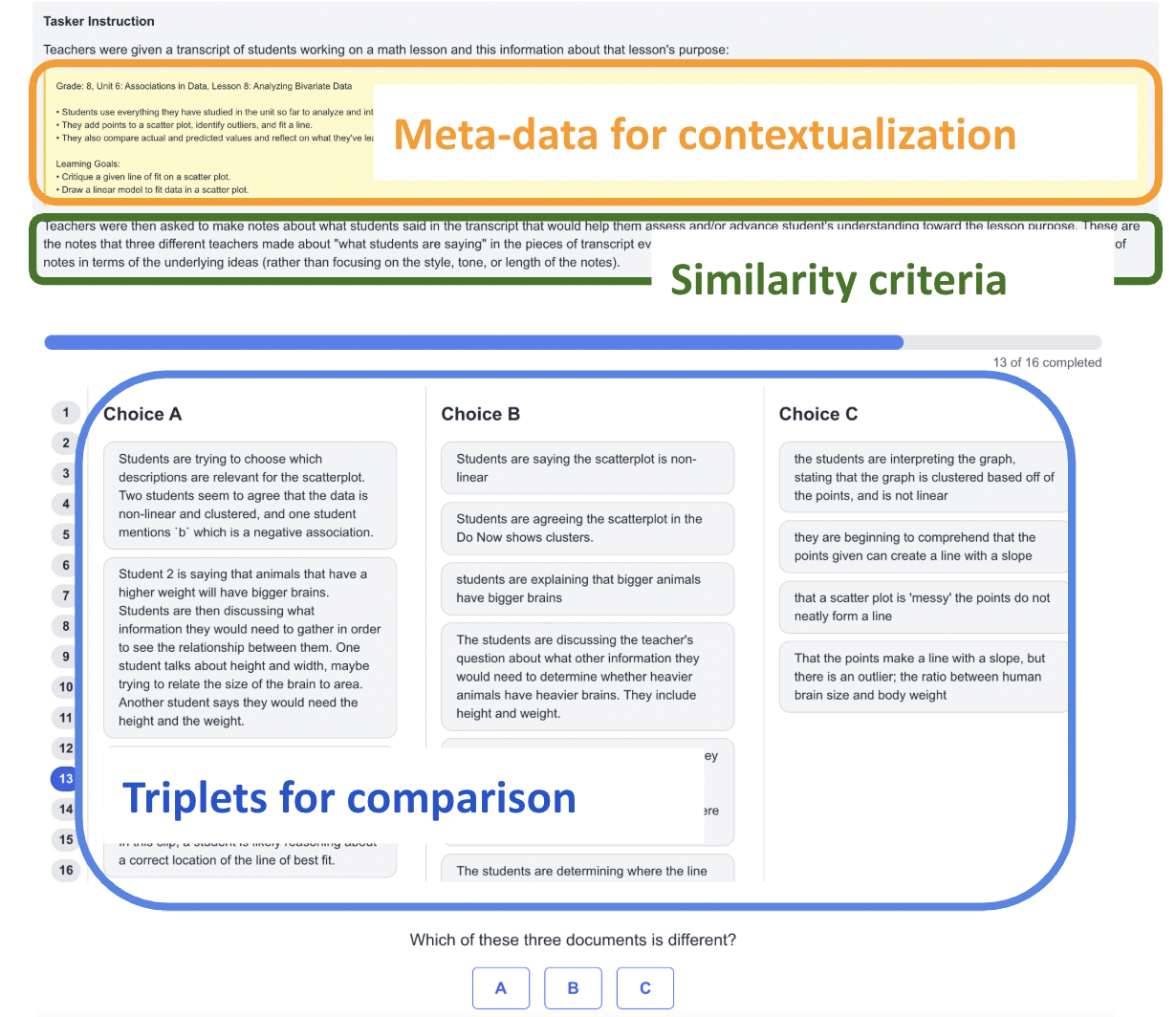}
  \caption{\small Evaluator interface for the odd-one-out task.}
  \label{fig:task_ui}
\end{figure}

We ensured that the similarity criteria being used were meaningful and grounded in educational practices by running a co-design process with 2 experts for the math reasoning and 4 experts for the feedback data. Through this process, we iterated on the similarity definition and evaluator instructions using 3-6 task samples. This led to the task guidelines and UI in Figure~\ref{fig:task_ui}. The template includes a section for metadata about the source text and expert-guided similarity criteria.

\subsection{Automated similarity measures} Expert judgments are the gold standard for comparing annotations, but they are expensive to collect and therefore difficult to scale. In contrast, automated similarity methods offer a scalable alternative, but they require validation to ensure agreement with expert judgments. To this end, we evaluate 36 automated methods falling into one of four approaches: lexical matching \raisebox{-0.2\height}{\includegraphics[height=1em]{lexical_clipart.png}};
     cosine similarity of text embeddings \raisebox{-0.2\height}{\includegraphics[height=1em]{embedding_clipart.png}};
    divergences of topic distributions \raisebox{-0.2\height}{\includegraphics[height=1em]{topic_clipart.png}}, and
     LLM-as-a-judge  \raisebox{-0.2\height}{\includegraphics[height=1em]{llm_clipart.png}}.

\paragraph{Lexical matching:} BLEU~\citep{papineni-etal-2002-bleu} is a widely used metric for lexical similarity based on n-grams. Since BLEU is asymmetric (measuring how much of a prediction overlaps with a target), we convert this to a symmetric metric between annotations $A$ and $B$ as follows: \begin{equation}
\frac{\text{BLEU}(A, B) + \text{BLEU}(B, A)}{2}.
\end{equation} We use a maximum n-gram order of 4.

\paragraph{Cosine similarity of text embeddings:} Texts can be embedded as vectors, and their similarity is measured using cosine distance. We examine 16 different text embedding models (Table \ref{tab:embedding_model_list}). For models with limited context windows, we split annotations into sentences and average sentence embeddings. 
 We also implement PCA-based post-processing of the text embeddings proposed by \citet{mu2018allbutthetopsimpleeffectivepostprocessing} which removes projections of the top principal components from the text embeddings.

\begin{table}[ht]
\centering
\begin{tabular}{p{\columnwidth}}
\small\textbf{Text embedding model} \\
\hline
\small Sentence-BERT~\citep{reimers2019sentencebertsentenceembeddingsusing} \\
\small MiniLM L6, 12~\citep{wang2020minilmdeepselfattentiondistillation} \\ 
\small MPNet~\cite{song2020mpnet} \\ 
\small SimCSE~\cite{gao2021simcse} \\ 
\small GTR-T5 (large, xl)~\cite{ni2021largedualencodersgeneralizable} \\ 
\small ModernBERT~\cite{warner2024smarterbetterfasterlonger} \\ 
\small E5-Base~\cite{wang2024textembeddingsweaklysupervisedcontrastive} \\ 
\small GTE (base, large)~\citep{li2023generaltextembeddingsmultistage} \\ 
\small GPT2 (large, xl)~\citep{Radford2019LanguageMA} \\ 
\small cpt-text (small, large)~\citep{neelakantan2022textcodeembeddingscontrastive} \\ 
\small ada~\citep{openaiada2022} \\
\end{tabular}
\caption{\small \textbf{List of text embedding models used for evaluation.} Some models, like GPT2, are from the same family but have different number of parameters.}
\label{tab:embedding_model_list}
\end{table}

\paragraph{Divergences of topic distributions:} Texts can be represented by topic distributions found using algorithms like BERTopic~\citep{grootendorst2022bertopic} and LDA~\citep{NIPS2001_296472c9}. Pairwise dissimilarity then becomes equivalent to measuring divergence between topic distributions of annotation pairs, $T_A$ and $T_B$. We use Hellinger distance as a divergence metric because it can allow for cases where some topics appear in only one of the annotations. We define similarity as one minus the distance:
\begin{equation}
\text{sim}(A, B) = 1 - \frac{1}{\sqrt{2}}\sqrt{\sum_{i=1}^{\text{$k$ topics}} (\sqrt{T_A[i]} - \sqrt{T_B[i]})^2}
\end{equation} Our hyperparameter search across different numbers of topics $k$ shows that $k$ affects agreement because it controls whether representations are too coarse (small $k$) or too fine-grained (large $k$). Specifically, we searched through minimum cluster sizes $2-15$ in BERTopic (HDBSCAN) and $k$ topic numbers through $(50, 75, 100, 125, 150)$ in LDA, and reported the best result.\footnote{BERTopic is implemented with open source packages \href{https://maartengr.github.io/BERTopic/getting_started/quickstart/quickstart.html}{BERTopic}, \href{https://umap-learn.readthedocs.io/en/latest/}{UMAP} and \href{https://pypi.org/project/hdbscan/}{HDBSCAN}, and LDA is implemented by \href{https://github.com/maria-antoniak/little-mallet-wrapper}{maria-antoniak/little-mallet-wrapper} on top of MALLET~\citep{McCallumMALLET}.}

\paragraph{LLM-as-a-judge for similarity:} Unlike previous approaches that output pairwise similarity scores between annotations $A$ and $B$, LLM-as-a-judge~\citep{zheng2023judgingllmasajudgemtbenchchatbot} can directly answer the odd-one-out tasks. The model is prompted with the same triplet multiple times ($n=5$), so each item's probability of being selected can be aggregated from multiple model answers. We test this method with different model types (GPT-4.1~\cite{openai2024gpt4technicalreport} and Claude-Sonnet-4~\cite{C3sonnet4}), temperatures (0.2 and 0.8), and prompt templates (with or without additional domain information). We used the prompt below for the feedback data (reasoning prompt in Appendix \ref{appendix:prompts}):

\begin{tcolorbox}[colback=yellow!10, colframe=black, coltitle=black, colbacktitle=white, title=\emph{Odd-One-Out} selection with LLM-as-a-Judge., fonttitle=\small\bfseries]
\label{prompt}
\small You are a helpful high school English Language Arts teaching assistant. The student was instructed to write an essay to the prompt: \texttt{Include meta data about the essay topic}, and three teachers were asked to give the student feedback for revising their writing. Select which feedback is most different in terms of what it asks the student to change in their writing. Focus on the feedback content, rather than judging based on the delivery (e.g., teacher's writing style, tone, length, use of questions versus directives in the feedback). Respond with \#\#A\#\# if feedback A is most different, \#\#B\#\# if feedback B is most different, and \#\#C\#\# if feedback C is most different.
\end{tcolorbox}

We also experimented with prompting LLMs for direct (pairwise) similarity scores, and found that this approach performs worse than prompting for the odd-one-out tasks (Appendix \ref{appendix:triplet-versus-direct}), so we keep only the triplet task setup for our main experiments.

\subsection{Evaluation of model similarity judgments with expert benchmarks} Since most automated similarity methods, except for LLM-as-a-judge, cannot directly pick the odd-one-out, we transform the pairwise similarity scores produced by the model into the probability of each item in a triplet being the most different. We assume that the probability of an item not being selected as the odd one out is proportional to that item's similarity with the other two items in the triplet. This leads to the model-predicted distribution that represents how likely each item $x \in \{A, B, C\}$ is to be selected:
\begin{equation}\label{eq:model_chosen}
\widehat P(x|A,B,C) = 1 - \frac{\sum_{j \not= x, j \in \{A, B, C\}} \text{sim}(x, j)}{\sum_{i \not= j, i, j \in \{A, B, C\}} \text{sim}(i, j)}.
\end{equation} This is compared to the human empirical distribution from Eq. \ref{eq:human_chosen} using Hellinger distance. Hellinger distance ranges from 0 (perfect agreement) to 1. Our experiment results report the average distance between the model predictions and the expert answers across all triplets in the evaluation dataset.

We chose Hellinger distance over other metrics, such as KL divergence, because KL asymmetrically penalizes under-estimation more than over-estimation, leading to counter-intuitive results. For example, given expert distribution $[0.95, 0.05, 0]$, KL divergence is lower for model prediction $[\frac{1}{3}, \frac{1}{3}, \frac{1}{3}]$ than for $[1, 0, 0]$, even though the latter correctly identifies the most dissimilar item as experts and therefore we would expect this to get a smaller penalty than uniform guessing.

\section{Datasets}
We applied \methodname{} to two educational annotation tasks: assessing student mathematical reasoning and feedback generation. 

\textbf{Student mathematical reasoning}. Eleven expert math teachers and four models each annotated four deidentified transcripts\footnote{These transcripts were collected from middle schools in a large urban public school district in California under an IRB for human subject research \#75294.} (ChatGPT and Claude Sonnet 3.5 each generated two annotations per transcript) of 6th-8th grade math lessons with inline comments. These comments analyzed students' mathematical reasoning, focusing on how specific pieces of student dialogue revealed their understanding of the key mathematical concepts. Annotations averaged 139.38 words (maximum: 556 words). Ten expert math educators (with more than 8 years of experience) then completed a total of 640 odd-one-out tasks, evaluating similarity based on the annotator's assessment of the students' understanding about the lesson's goals. 

\textbf{Essay feedback.} We used 18 student essays from open-source datasets~\citep{hamner2012asap}, with inline feedback from 32 experienced English Language Arts (ELA) educators~\citep{mah5213040sentence}. We additionally included four LLM-generated feedback per student essay (ChatGPT, Gemini 2.5 Pro, Claude Sonnet 3.5, and DeepSeek V2.5). Annotations averaged 142.98 words (maximum: 477 words). These comments focused on various aspects of student writing that needed revision (e.g., grammar, word choice, and organization). Seven ELA experts (one with 6 years of teaching and the rest with more than 8 years of experience) completed a total of 458 odd-one-out tasks, evaluating similarity based on the types of revisions requested by the teachers. 

Expert evaluators on both domains were compensated at an hourly rate of \$50, and each contributed between 3-5 hours to the odd-one-out tasks. Further data collection details are in Appendix ~\ref{appendix:data_collection}.

\section{Results}

Using these 36 similarity measures (including 16 text embeddings from Table \ref{tab:embedding_model_list} and their post-processing variants) and two datasets, we quantify consistency across the measures (RQ1). Next, evaluate similarity measures against expert judgments (RQ2), and probe when/why they diverge by looking at task difficulty, style sensitivity, and lexical overlap (RQ3). Finally, we assess how the best-performing method can triage human work through outlier detection and annotator-pair ranking (RQ4).

\paragraph{RQ 1: \emph{How much do automated similarity measures correlate with each other?}} 

We computed Pearson correlations between pairwise similarities from (i) BLEU (4-gram), (ii) 16 text-embedding models, (iii) topic distributions (LDA, $k$=100 topics), and (iv) LLM-as-a-judge (prompting for both triplet judgments and continuous pairwise scores). While the triplet judgment setup (\emph{LLM-as-a-judge triplet}) is our preferred specification for LLM-as-a-judge, here we also include a variant where GPT-4.1 is prompted to produce continuous pairwise similarity scores for comparability (\emph{LLM-as-a-judge direct}); we sampled five outputs per pair and averaged them.

\begin{figure}[htbp]
  \includegraphics[width=\columnwidth]{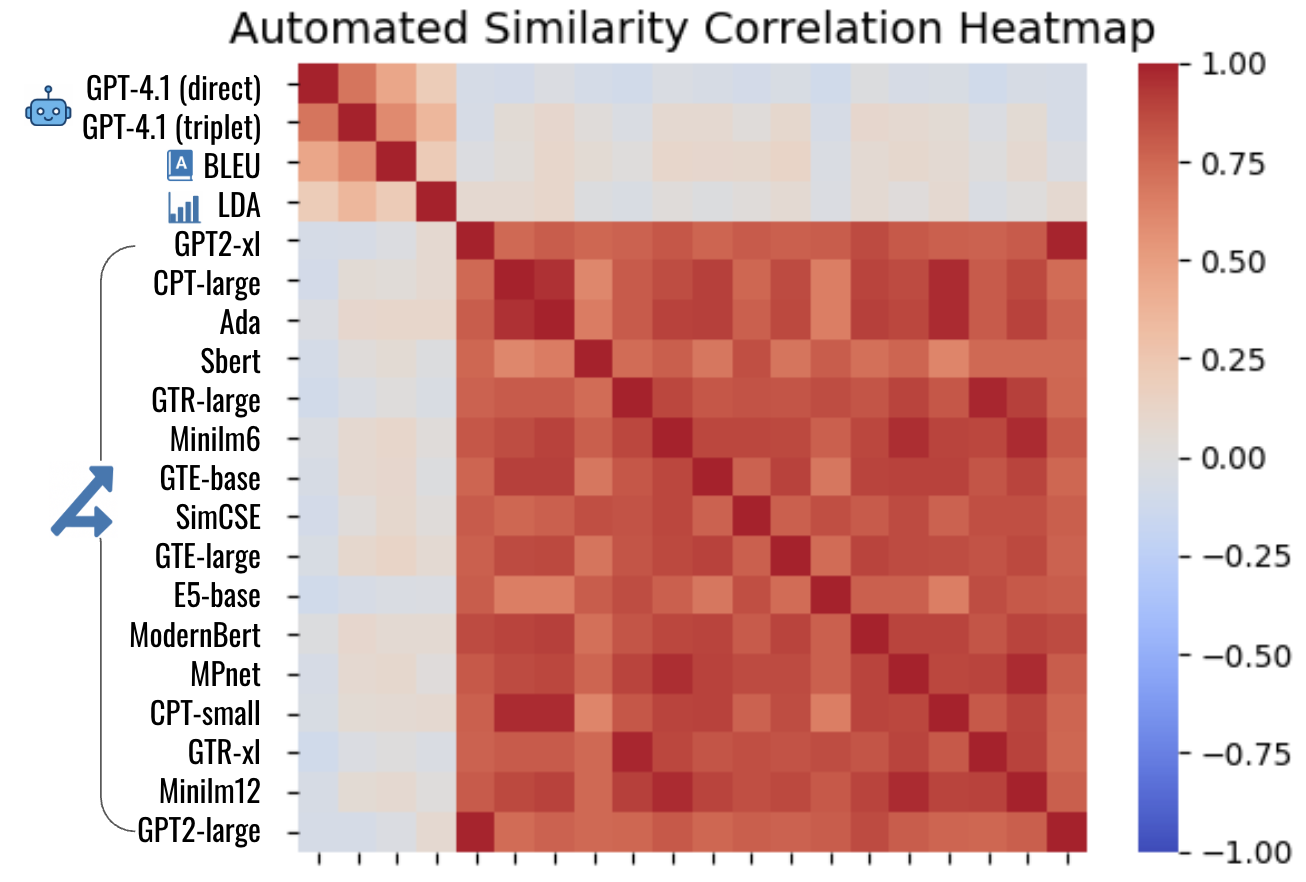}
  \caption{\small Pearson correlation coefficients of different similarity measures on the reasoning data. Cross-family pairs show no to weak correlation, while correlations within embedding-based measures are higher.}\label{fig:math_correlation}
\end{figure}

On the reasoning data (Fig.~\ref{fig:math_correlation}), 56 out of 190 cross-method pairs show non-significant correlations (based on $p < 0.05$). Embedding-based measures show moderate to high correlations with one another, but they do not correlate significantly with BLEU, LDA, and both variants of LLM-as-a-judge. LDA also correlates weakly with the LLM-as-a-judge. LLM-as-a-judge direct and triplet correlate strongly with each other. On the feedback data, correlations of cross-method pairs are generally higher (Appendix~\ref{appendix:essay_correlation}).

Inconsistency among measures means that they can lead to conflicting conclusions about which annotations are ``similar." This highlights the need to validate automated measures against human judgments and to do so \emph{by domain}, as reliability of different methods may vary across tasks and datasets.

\paragraph{RQ 2: \emph{How well do automated similarity methods align with expert similarity judgments?}}

We quantify alignment by comparing each method’s predictions on the odd-one-out triplets to the expert answer distributions using Hellinger distance (lower is better; 0 = perfect match). A uniform guess serves as a naïve baseline. As shown in Table~\ref{tab:rq1}, BLEU and embedding-based similarities perform near this baseline (especially on reasoning), while \emph{LLM-as-a-judge} is best in both domains, improving alignment by 18\% over alternatives in feedback and 11\% improvement in reasoning. This trend holds across model types (GPT-4.1, Claude Sonnet 4); lower temperature (0.2) outperforms 0.8; and surprisingly, adding extra education-related context does not help (Appendix~\ref{appendix:ablations}). Extensive embedding variants, with and without PCA post-processing, remain close to uniform on the reasoning dataset (Appendix~\ref{appendix:model_eval}).

\begin{table}[htbp]

  \centering
  \resizebox{\columnwidth}{!}{%
    \begin{tabular}{l | c | c}
      \toprule
       \textbf{Method}  & \textbf{Reasoning} & \textbf{Feedback} \\ \hline
      Uniform & 0.650 ($\pm$ 0.000) & 0.493 ($\pm$ 0.015) \\ 
      Lexical matching (BLEU) {\includegraphics[height=1em]{lexical_clipart.png}} & 0.650 ($\pm$ 0.000)  & 0.449 ($\pm$ 0.026)  \\
      Text embedding (gpt2-xl){\includegraphics[height=1em]{embedding_clipart.png}} & 0.653 ($\pm$ 0.009) &  0.481 ($\pm$ 0.004)\\
       Text embedding (ada){\includegraphics[height=1em]{embedding_clipart.png}} & 0.650 ($\pm$ 0.000) & 0.482 ($\pm$ 0.004) \\
      Topic distribution (LDA) {\includegraphics[height=1em]{topic_clipart.png}} & 0.639 ($\pm$ 0.002) & 0.476 ($\pm$ 0.015) \\
      Topic distribution (BERTopic) {\includegraphics[height=1em]{topic_clipart.png}} & 0.619 ($\pm$ 0.006) & 0.495 ($\pm$ 0.016) \\
      \textbf{LLM-as-judges} (GPT-4.1) {\includegraphics[height=1em]{llm_clipart.png}} & \textbf{0.541 ($\pm$ 0.006)} & \textbf{0.374 ($\pm$ 0.029)} \\
      \bottomrule
    \end{tabular}  
  }
  \caption{\small Average Hellinger distances between the model outputs and the expert answers to the odd-one-out tasks (640 expert labels from reasoning and 458 from feedback). Values range between 0 and 1, with 0 indicating perfect agreement. We report the mean and the standard error.}
\label{tab:rq1}
\end{table}

\paragraph{\emph{Does model performance vary by task difficulty?}} Some triplets are inherently more difficult for experts to identify the odd one out from. We captured task difficulty by collecting ratings of each triplet on a three-point scale: difficult (-1), neutral (0), and easy (1). We averaged these ratings across experts and compared model performance on the easy (mean rating $> 0$) and difficult (mean $< 0$) subsets. This yielded 38\% easy and 11\% difficult triplets for reasoning, and 65\% easy and 16\% difficult for feedback. We computed average Hellinger distances within each subset and tested for significant differences using two-tailed independent samples t-tests. Table \ref{tab:rq1_task_difficulty_dependency} shows that only LLM-as-a-judge achieves significantly better performance on the easy tasks consistently across two domains.

\begin{table}[h!]
\centering
\resizebox{\columnwidth}{!}{
\begin{tabular}{lcc}
\toprule
\textbf{Method} & \textbf{Reasoning (Easy)} & \textbf{Reasoning (Hard)} \\
\hline
Lexical matching (BLEU) {\includegraphics[height=1em]{lexical_clipart.png}} & 0.65 (0.65) & 0.65 (0.65) \\
Text embedding (gpt2-xl) {\includegraphics[height=1em]{embedding_clipart.png}} & 0.65 (0.65) & 0.65 (0.65) \\
Text embedding (ada) {\includegraphics[height=1em]{embedding_clipart.png}} & 0.65 (0.65) & 0.65 (0.65) \\
\textbf{Topic distribution (LDA)} {\includegraphics[height=1em]{topic_clipart.png}} & \textbf{0.62 (0.63)*} & \textbf{0.64 (0.64)*} \\
Topic distribution (BERTopic) {\includegraphics[height=1em]{topic_clipart.png}} & 0.58 (0.61) & 0.63 (0.65)  \\
\textbf{LLM-as-judges (GPT-4.1)} {\includegraphics[height=1em]{llm_clipart.png}} & \textbf{0.49 (0.39)**} & \textbf{0.58 (0.56)**} \\
\textbf{LLM-as-judges (Sonnet 4)} {\includegraphics[height=1em]{llm_clipart.png}} & \textbf{0.49 (0.39)**} & \textbf{0.61 (0.63)**} \\
\end{tabular}}

\resizebox{\columnwidth}{!}{
\begin{tabular}{lcc}
\toprule
\textbf{Method} & \textbf{Feedback (Easy)} & \textbf{Feedback (Hard)} \\
\hline
Lexical matching (BLEU) {\includegraphics[height=1em]{lexical_clipart.png}} & 0.42 (0.43) & 0.50 (0.54)  \\
Text embedding (gpt2-xl) {\includegraphics[height=1em]{embedding_clipart.png}} & 0.50 (0.48) & 0.46 (0.43) \\
Text embedding (ada) {\includegraphics[height=1em]{embedding_clipart.png}} & 0.51 (0.48) &  0.46 (0.43) \\
Topic distribution (LDA) {\includegraphics[height=1em]{topic_clipart.png}} & 0.49 (0.49) & 0.47 (0.44) \\
Topic distribution (BERTopic) {\includegraphics[height=1em]{topic_clipart.png}} & 0.49 (0.51) & 0.51 (0.47) \\
\textbf{LLM-as-judges (GPT-4.1)} {\includegraphics[height=1em]{llm_clipart.png}} & \textbf{0.36 (0.32)**} & \textbf{0.56 (0.55)**} \\
\textbf{LLM-as-judges (Sonnet 4)} {\includegraphics[height=1em]{llm_clipart.png}} & \textbf{0.35 (0.32)**} & \textbf{0.58 (0.55)**} \\
\hline
\end{tabular}}
\caption{\small Model performance on easy and difficult task subsets. We report the mean Hellinger distance (and the median in the parentheses) for each subset. P values, based on a two-tailed t-test: *$p<0.05$, **$p<0.01$. Only LLM-as-a-judge shows significant performance difference on both datasets.}\label{tab:rq1_task_difficulty_dependency}
\end{table} To provide a different lens on this gap, we also report model prediction accuracy: 1 if the model matches the majority expert answer, and 0 otherwise. GPT-4.1 as a judge improves from 44.1\% (hard) to 70.6\% (easy) in reasoning, and Claude Sonnet 4 from 39.7\% to 71.4\%. Similar trends are observed in feedback (Appendix \ref{appendix:llm-as-judge-performance}). Overall, these evaluation results show that LLM-as-a-judge aligns most closely with expert similarity judgments, \emph{especially when clear outliers exist,} but there remains considerable room for improvement for all automated methods.

\paragraph{RQ 3: \emph{What might cause models to disagree with experts?}} We hypothesize two possible factors contributing to model misalignment, especially for embedding-based and lexical similarities: (i) sensitivity to style, and (ii) lack of lexical overlap with the source text due to the interpretive nature of the annotations.

\textbf{Style.} We noticed a discrepancy between the annotations that embedding models identified as most dissimilar and those chosen by experts. We hypothesized that this was due to differences in writing style (formal vs casual). To test this, we took two pieces of annotations from the reasoning dataset with the lowest average similarity scores and rewrote them in a more formal style without changing their content (see Appendix \ref{appendix:style_examples} for changes). Table \ref{tab:rq2_style} shows that the embedding similarities increase by 20–85\% after this rewriting, whereas LLM-as-a-judge is more stable, showing only 2-5\% change.
\begin{table}[ht]
\centering
  \resizebox{\columnwidth}{!}{%
    \begin{tabular}{l | c }
      \toprule
       \textbf{Text embedding model {\includegraphics[height=1em]{embedding_clipart.png}}}  & \textbf{\% change in similarity scores} \\ \hline
       Sentence-BERT &  $17.35 \uparrow$ \\ 
  ModernBERT &  $23.45 \uparrow$\\ 
  GPT-XL  & $26.08 \uparrow$ \\ 
  MPNet & $68.16 \uparrow$\\ 
  MiniLM-L12 &  $82.30 \uparrow$ \\ 
  \textbf{LLM-as-a-judge {\includegraphics[height=1em]{llm_clipart.png}}}  & \textbf{\% change in similarity scores} \\ \hline
  GPT-4.1   &   $5.00 \uparrow$ \\ 
  Claude Sonnet 4 & $2.03 \downarrow$ \\ \bottomrule
    \end{tabular}%
    }
\caption{We report the percentage difference (\%) in similarity scores of different models after the style manipulation. Robust models should not result in dramatic score changes, because we controlled the content to be the same.}
\label{tab:rq2_style}
\end{table}

\textbf{Lack of grounding in the source text.} Table \ref{tab:rq1} shows that BLEU's average Hellinger distances are close to uniform guessing. BLEU is uninformative when annotations have little to no lexical overlap.  This is expected for open-ended, interpretive tasks, where annotators (teachers) paraphrase rather than directly quote the student text. BLEU scores between annotations and source texts are extremely low: 0.0127 (SD 0.022) for feedback and 0.0004 (SD 0.002) for reasoning. Given this little overlap with the source text, low BLEU scores among annotations are unsurprising. While BLEU is useful for summarization and translation, where lexical similarity to a target is desirable and expected, it performs poorly for tasks that involve interpreting and paraphrasing from one text form (e.g., student dialogue or essays) to another (e.g., teacher's assessment of student reasoning or feedback comments).


\paragraph{RQ 4: \emph{What insights do automated similarity methods provide that can reduce human workload?}} The goal of the previous section was to evaluate and select reliable automated similarity methods for comparing LLM interpretations to domain experts. While agreement with expert judgments can be improved across all methods, LLM-as-a-judge may still be useful for answering some data-driven questions that would otherwise require manual human evaluation. We explore this potential through two tasks: outlier annotation detection and ranking annotator pairs by similarity.
\begin{table}[htbp]

  \centering
  \resizebox{\columnwidth}{!}{%
    \begin{tabular}{l | c | c}
      \toprule
       \textbf{LLM-as-judges  {\includegraphics[height=1em]{llm_clipart.png}}}  & \textbf{Reasoning \%} & \textbf{Feedback \%} \\ \hline
      GPT-4.1 & 86.67 & 61.90 \\
      Claude 4 Sonnet & 80.00 & 71.43 \\
      \bottomrule
    \end{tabular}  
  }
  \caption{\small Outlier detection rate by LLM-as-a-judge using the top quartile of the expert-selected odd ones as the true outliers.}
\label{tab:rq4-outlier}
\end{table}

\paragraph{Outlier detection.} One natural question to ask about LLM and human-generated annotations is whether there exist clear outliers. We define outliers as annotations that had the highest possibility of being selected as the odd-one across different triplets. We used the top quartile from the expert data to establish the outlier thresholds and assessed whether the model's top quartile could detect the same outliers. Table \ref{tab:rq4-outlier} shows that GPT-4.1 and Sonnet 4 detect over 80\% of the outliers in reasoning and over 60\% in feedback. 

\begin{table}[htbp]

  \centering
  \resizebox{\columnwidth}{!}{%
    \begin{tabular}{l | c | c}
      \toprule
       \textbf{LLM-as-judges  {\includegraphics[height=1em]{llm_clipart.png}}}  & \textbf{Reasoning $\rho$} & \textbf{Feedback $\rho$} \\ \hline
      GPT-4.1 & 0.683** & 0.618**  \\
      Claude Sonnet 4 & 0.589**  & 0.613**  \\
      \bottomrule
    \end{tabular}  
  }
  \caption{\small Spearman rank correlation $\rho$ between LLM and expert produced rankings of pairwise similarities. **: $p<0.01$. We convert the answers to the odd-one-out tasks by LLMs and human experts into pairwise similarity scores by up-voting pairs that appeared in the same triplet but were not chosen (higher similarity) and down-voting pairs where either one was chosen (lower similarity). We generate rankings of the most and least similar pairs according to these pairwise scores and correlate the rankings generated by LLMs vs humans.}
\label{tab:rq4-rankings}
\end{table}

\paragraph{Ranking annotator pairs.} Researchers may be interested in comparing annotator pairs (LLM-LLM, human-human, human-LLM) to see whether LLMs are more similar to each other than to humans, and whether human-to-human variability is greater than differences among LLMs. Doing so requires ranking all annotator pairs, which is costly for humans to evaluate because the number of pairs grows as $n$ choose 2. As an alternative, LLM-as-a-judge can be used to generate pairwise similarity rankings. We demonstrate this by converting odd-one-out task responses into relative similarity scores and ranking annotator pairs using these derived scores. We repeat the same steps with expert relative judgments and compare the resulting LLM and expert rankings using Spearman rank correlation. Table \ref{tab:rq4-rankings} shows that LLM-as-a-judge achieves moderately high correlations with experts ($\rho: 0.58-0.68$, and $p < 0.01$) on both datasets. These results demonstrate that automated similarity methods can support scalable data exploration (e.g., identifying outliers and ranking similar annotator pairs) providing insights that would otherwise require costly human evaluation.



\section{Discussion \& Conclusion}
As interest grows in evaluating LLM outputs against experts' for open-ended tasks~\cite{spangher-etal-2024-llms, Mu_oz_Ortiz_2024, ismayilzada2025evaluatingcreativeshortstory}, we need evaluation that goes beyond surface features like word overlap or style, and capture alignment of underlying ideas. Our evaluation shows that most existing similarity measures show poor agreement with experts. In contrast, LLM-as-a-judge achieves 11-18\% improvement in agreement with experts than lexical and vector-based methods, and performs especially well when clear outliers exist. Among the models evaluated, we conclude that prompting LLMs for relative judgments offers the most promising path for automated idea-level similarity evaluation, but we recommend domain-specific validation against human benchmarks. For NLP and social science researchers, unstructured data exploration, such as clustering similar ideas and comparing shared themes, is crucial but laborious and time-intensive. Automated similarity methods can reduce human workload and support large-scale data analysis by highlighting similarities and differences of ideas in unstructured annotations. We encourage researchers from different domains to consider applying \methodname{} and other validation approaches to domains beyond education.

\textbf{Future direction.} An interesting future direction to explore is whether similarity metrics/models behave differently when judging LLM-generated versus human-generated text pairs. For example, it would be worthwhile to investigate if models are more or less effective at judging similarity in ideas generated by LLMs (either from the same or different model families) than at comparing human-generated pairs. Furthermore, applying our evaluation framework to LLM-generated text pairs can help provide insights into models' output homogeneity highlighted in recent work ~\citep{jiang2025artificialhivemindopenendedhomogeneity, zhang2025verbalizedsamplingmitigatemode}.

\section*{Limitations} 
\textbf{Challenge of directly quantifying pairwise similarity of ideas.} While it would be ideal to collect pairwise similarity scores from humans given a text pair, doing so reliably at scale is difficult. We believe that there is no well-established method for humans to estimate the "similarity of ideas behind texts," and propose it as a crucial challenge as open-ended evaluations of LLM-generated texts become increasingly important and popular. Instead of collecting pairwise similarities from humans, this work proposes "odd-one-out" evaluation as an alternative to reduce human cognitive load and increase consistency. While this differs theoretically from absolute pairwise distance, it is a relatively easy and intuitive task for humans to perform. We demonstrate how to build an evaluation protocol for models and existing semantic similarity measures that can match human outputs. Our insight is to compare humans and models based on how well the models' similarity estimates match humans’ triplet answer distributions.

\textbf{Human guidance \& validation.} Defining dimensions of similarity with domain experts is crucial to our analysis. For instance, even within math education, the criteria for measuring similarity may vary depending on the downstream task and the data from which the annotations are generated. Therefore, collaborating with human experts is essential, though it can be expensive compared to fully automated evaluation pipelines. 

\textbf{Focus of our work on education.} In this paper, we focus on two educational tasks, assessing student math reasoning and providing essay feedback, but there are many other tasks, both within and beyond education that also require domain expertise to generate meaningful interpretations and measure similarity. While our general evaluation framework can be applied to various open-ended, interpretive text-generation tasks, we recommend that validating with domain-specific expert oversight is crucial.

\textbf{Challenges of aligning LLMs with domain experts.} While LLM-as-a-judge shows promise compared to other similarity measures, our experimental results suggest that they still lack domain expertise appropriate for full automation, and prior work has questioned the reliability of their outputs~\cite{feuer2025style}. Therefore, we recommend careful validation of all methods discussed in our work, including LLM-as-a-judge, before applying them to any new dataset.

\section*{Ethical considerations} Math classroom transcript data was collected under an IRB for human subject research (approval number: 75294) in which teachers, students, and student guardians consented to their participation. The feedback data was collected in prior work \cite{mah5213040sentence, tan2024reframing}. Recruited and compensated experts for similarity assessments only used deidentified data.

Regarding the intended use of our method, we propose a benchmarking task for measuring the similarity of ideas as a scalable way to assess LLM-generated responses against those of expert humans. However, high similarity scores do not imply endorsing indiscriminate use of LLMs in education. Other social and educational risks of using LLMs in classrooms or to interact with students should be considered holistically, regardless of their idea-level similarity to human teachers.

\clearpage
\bibliography{custom}
\appendix
\clearpage
\section{Prompting LLM-as-a-judge with relative judgments versus direct score assignment.}\label{appendix:triplet-versus-direct}
\begin{table}[h!]
\centering
\resizebox{\columnwidth}{!}{
\begin{tabular}{lcc}
\toprule
\textbf{Method} & Reasoning & Feedback \\
Direct & 0.624 (0.003) & 0.452 (0.015) \\ \hline
\textbf{Triplet} & \textbf{0.541 (0.006)} &  \textbf{0.374 (0.029)}\\
\hline
\end{tabular}}
\caption{We added a variant of prompting LLMs for direct similarity scores (\emph{direct}) rather than answering the relative judgment tasks (\emph{triplet}). We sampled five outputs per pair to average and converted the pairwise similarity scores into the probability of each item in a triplet being selected as the most different, following the steps discussed in Section 3.3. We report the mean and the standard error across triplets. Smaller Hellinger distances (closer to 0) indicate better agreement with experts.}
\end{table}

\section{Correlations of different automated similarity measures with each other.}

\begin{figure}[htbp]
  \includegraphics[width=\columnwidth]{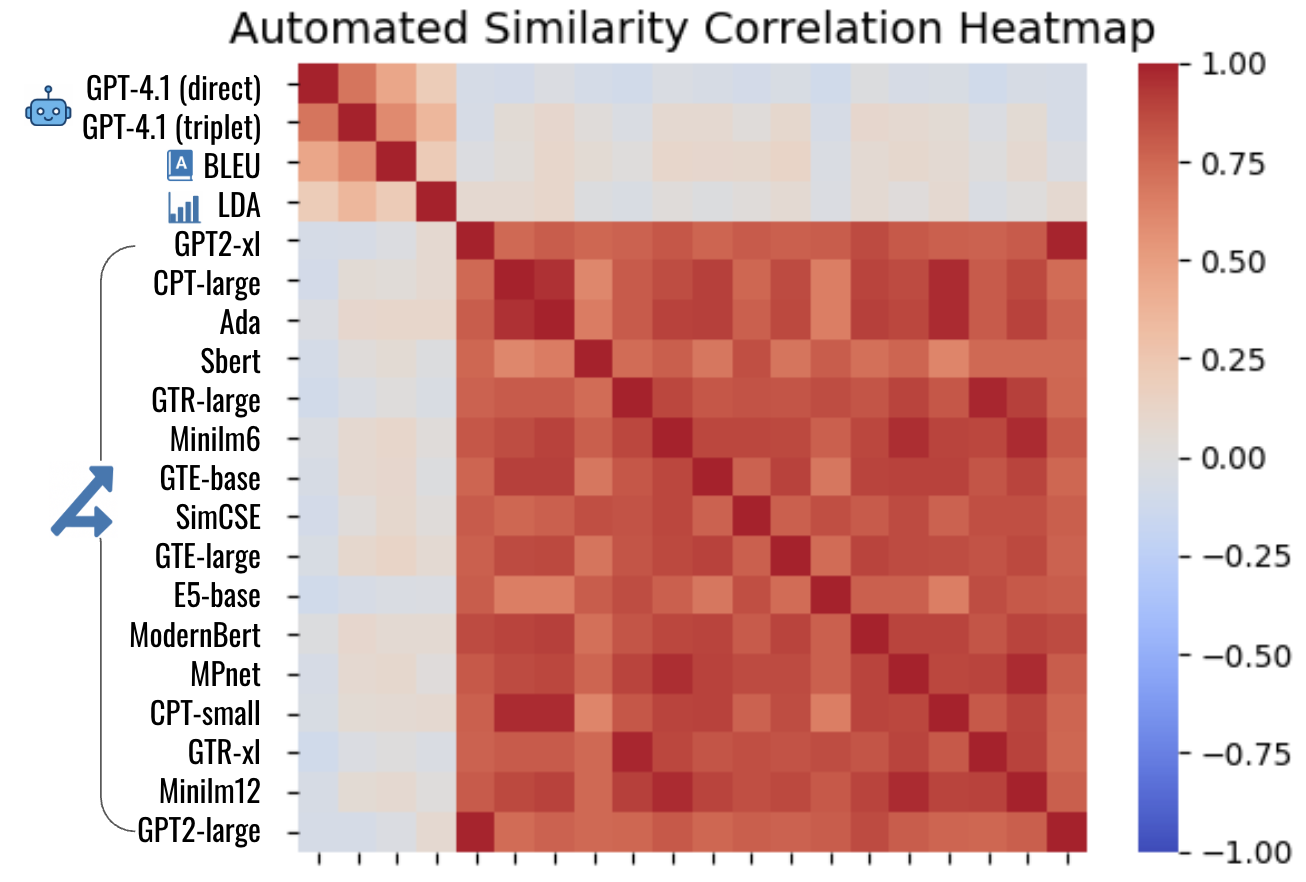}
  \caption{\small \textbf{Pearson correlation coefficients $\rho$ of different similarity methods / models in math reasoning.} Negative correlations suggest that different metrics may yield different conclusions about whether the annotations are similar or not. Under one metric, the same pair may be considered to have high similarity score, but under another metric, considered to have low similarity. Pairwise scores are computed for the math data using different similarity models / methods and their correlations are computed with Pearson.}
  \label{appendix:math_correlation}
\end{figure}

\begin{figure}[htbp]
  \includegraphics[width=\columnwidth]{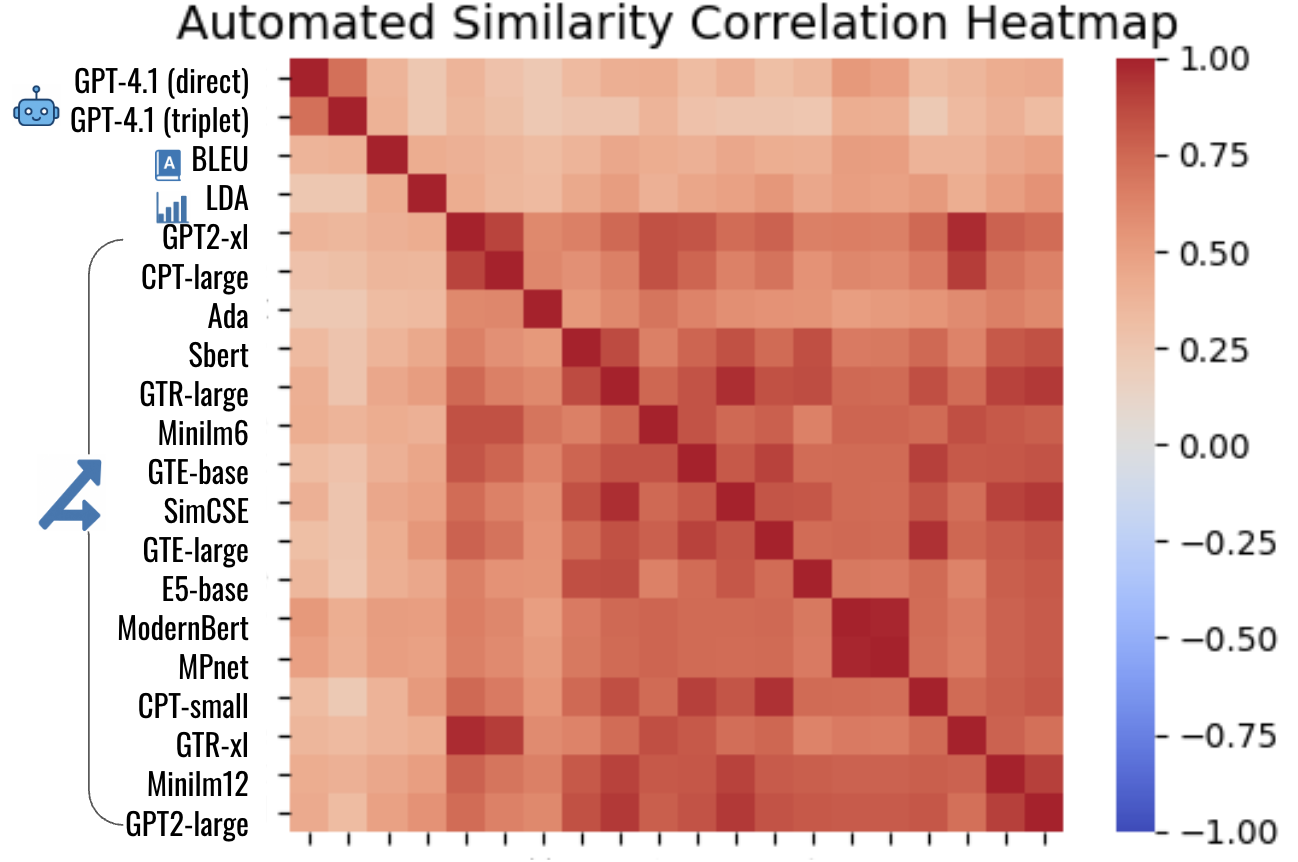}
  \caption{\small \textbf{Pearson correlation coefficients $\rho$ of different similarity methods / models in essay feedback.} Negative correlations suggest that different metrics may yield different conclusions about whether the annotations are similar or not. Under one metric, the same pair may be considered to have high similarity score, but under another metric, considered to have low similarity. Pairwise scores are computed for the math data using different similarity models / methods and their correlations are computed with Pearson.}
  \label{appendix:essay_correlation}
\end{figure}

\section{Text embedding models used in evaluation.} See Table \ref{tab:embeddings}.
\begin{table}[!h]
  \resizebox{\columnwidth}{!}{
\centering
\begin{tabular}{lccc}
\small\textbf{Text embedding model} & \small\textbf{Context length} & \small\textbf{Embedding dim} & \small\textbf{Open-sourced?} \\
\hline
\small Sentence-BERT~\citep{reimers2019sentencebertsentenceembeddingsusing}  & \small 128 & \small 768 & \cmark \\ \hline
\small MiniLM L6, 12~\citep{wang2020minilmdeepselfattentiondistillation} & \small 256 & \small 384  & \cmark \\ \hline 
\small MPNet~\cite{song2020mpnet} & \small 382 & \small 768 & \cmark \\ 
\hline
\small SimCSE~\cite{gao2021simcse} & \small 512 & \small 1024  & \cmark \\ \hline
\small GTR-T5 (large, xl)~\cite{ni2021largedualencodersgeneralizable} & \small 512 & \small 768 & \cmark \\ \hline 
\small ModernBERT~\cite{warner2024smarterbetterfasterlonger} & \small 8192 & \small 1024 & \cmark \\ \hline 
\small E5-Base~\cite{wang2024textembeddingsweaklysupervisedcontrastive} & \small 512 & \small 768 & \cmark \\ \hline 
\small GTE (base, large)~\citep{li2023generaltextembeddingsmultistage} & \small 8192 & \small 768, 1024 & \cmark \\ \hline 
\small GPT2 (large, xl)~\citep{Radford2019LanguageMA} & \small 1024 & \small 1280, 1600 & \cmark \\ 
\hline \small cpt-text (small, large)~\citep{neelakantan2022textcodeembeddingscontrastive} & \small 8192 & \small 1536, 3072 & \xmark \\ \hline 
\small ada~\citep{openaiada2022} & \small 8192 & \small 1536 & \xmark \\
\end{tabular}}
\caption{These are the text embedding models we used in our experiments to compute cosine similarity scores. We evaluated a total of 16 available models, both open-sourced and proprietary. For models with a maximum context length of less than 1024 tokens, we embedded each sentence in the document and used the average of the sentence embeddings to represent the document. If the model allowed longer contexts, we embedded the full document.}
\label{tab:embeddings}
\end{table}

\section{Evaluation of text embedding models against expert similarity judgments}\label{appendix:model_eval} See Table \ref{tab:embedding_reasoning} and \ref{tab:embedding_feedback}.

\begin{table}[!h]
\centering
\resizebox{\columnwidth}{!}{
\begin{tabular}{l | cc}
    \toprule
    \textbf{Embedding Model} & \textbf{Original} & \textbf{Post-processing by} \citet{mu2018allbutthetopsimpleeffectivepostprocessing} \\
    \hline
    gtr-large      & 0.651 ($\pm$ 0.006) & 0.688 ($\pm$ 0.009)  \\
    gtr-xl         & 0.652 ($\pm$ 0.007) & 0.696 ($\pm$ 0.009)  \\
    e5-base        & 0.650 ($\pm$ 0.002) & 0.678 ($\pm$ 0.009)  \\
    sbert          & 0.652 ($\pm$ 0.002) & 0.682 ($\pm$ 0.010) ) \\
    gte-base       & 0.651 ($\pm$ 0.002) & 0.669 ($\pm$ 0.009)  \\
    gte-large      & 0.650 ($\pm$ 0.002) & 0.684 ($\pm$ 0.009)  \\
    simcse         & 0.651 ($\pm$ 0.001) & 0.681 ($\pm$ 0.010) \\
    mpnet          & 0.650 ($\pm$ 0.002) & 0.676 ($\pm$ 0.009)  \\
    mini-l6        & 0.652 ($\pm$ 0.006) & 0.665 ($\pm$ 0.009)  \\
    gpt2-large     & 0.653 ($\pm$ 0.009) & 0.673 ($\pm$ 0.010)  \\
    gpt2-xl        & 0.653 ($\pm$ 0.009) & 0.699 ($\pm$ 0.009) \\
    mini-l12       & 0.652 ($\pm$ 0.003) & 0.669 ($\pm$ 0.009)  \\
    modern-bert    & 0.650 ($\pm$ 0.001) & 0.667 ($\pm$ 0.009)  \\
    cpt-text-small & 0.651 ($\pm 0.002$) & 0.687 ($\pm$ 0.010)  \\
    cpt-text-large & 0.652 ($\pm 0.002$) & 0.669 ($\pm$ 0.010)\\
    ada   & 0.650 ($\pm 0.000$) & 0.672 ($\pm$ 0.010) \\
    \bottomrule
\end{tabular}}
\caption{\small \textbf{Average Hellinger distances} between the model outputs and the expert answers to the odd-one-out tasks (640 expert labels from the reasoning dataset). Values range between 0 and 1, with 0 indicating perfect agreement. We report the mean and the standard deviation across the triplets.}\label{tab:embedding_reasoning}
\end{table}

\begin{table}[!h]
\centering
\resizebox{\columnwidth}{!}{
\begin{tabular}{l | ccc}
    \toprule
    \textbf{Embedding Model} & \textbf{Original} & Post-processing by \citet{mu2018allbutthetopsimpleeffectivepostprocessing} & Post-processing by ~\citet{raunak-etal-2020-dimensional} \\
    \hline
    gtr-large      & 0.481 ($\pm$ 0.004) & 0.529 ($\pm$ 0.024) & 0.440 ($\pm$ 0.017) \\
    gtr-xl         & 0.489 ($\pm$ 0.015) & 0.563 ($\pm$ 0.021) & 0.456 ($\pm$ 0.016) \\
    e5-base        & 0.490 ($\pm$ 0.011) & 0.586 ($\pm$ 0.022) & 0.449 ($\pm$ 0.018) \\
    sbert          & 0.489 ($\pm$ 0.009) & 0.552 ($\pm$ 0.022) & 0.480 ($\pm$ 0.022) \\
    gte-base       & 0.481 ($\pm$ 0.005) & 0.532 ($\pm$ 0.022) & 0.468 ($\pm$ 0.016) \\
    gte-large      & 0.488 ($\pm$ 0.007) & 0.522 ($\pm$ 0.021) & 0.469 ($\pm$ 0.015) \\
    simcse         & 0.485 ($\pm$ 0.006) & 0.498 ($\pm$ 0.022) & 0.415 ($\pm$ 0.018) \\
    mpnet          & 0.484 ($\pm$ 0.006) & 0.510 ($\pm$ 0.021) & 0.453 ($\pm$ 0.017) \\
    mini-l6        & 0.482 ($\pm$ 0.005) & 0.552 ($\pm$ 0.021) & 0.463 ($\pm$ 0.017) \\
    gpt2-large     & 0.482 ($\pm$ 0.004) & 0.545 ($\pm$ 0.023) & 0.416 ($\pm$ 0.019) \\
    gpt2-xl        & 0.481 ($\pm$ 0.004) & 0.511 ($\pm$ 0.022) & 0.430 ($\pm$ 0.020) \\
    mini-l12       & 0.481 ($\pm$ 0.004) & 0.540 ($\pm$ 0.020) & 0.453 ($\pm$ 0.018) \\
    modern-bert    & 0.481 ($\pm$ 0.004) & 0.517 ($\pm$ 0.024) & 0.432 ($\pm$ 0.017) \\
    cpt-text-small & 0.485 ($\pm$ 0.007) & 0.492 ($\pm$ 0.022) & 0.447 ($\pm$ 0.014) \\
    cpt-text-large & 0.482 ($\pm$ 0.005) & 0.525 ($\pm$ 0.017) & 0.445 ($\pm$ 0.015) \\
    ada            & 0.482 ($\pm$ 0.004) & 0.485 ($\pm$ 0.022) & 0.425 ($\pm$ 0.016) \\
    \bottomrule
\end{tabular}}
\caption{\small \textbf{Average Hellinger distances} between the model outputs and the expert answers to the odd-one-out tasks (458 from the feedback datasets). Values range between 0 and 1, with 0 indicating perfect agreement. We report the mean and the standard deviation across the triplets.}\label{tab:embedding_feedback}
\end{table}

\section{Ablations with different prompting hyperparameters.}\label{appendix:ablations} See Table \ref{appendix:prompting-parameters}.
\begin{table}[h!]
\centering
\resizebox{\columnwidth}{!}{
\begin{tabular}{lcc}
\toprule
\textbf{Temperature} & \textbf{Low (0.2)} & \textbf{High (0.8)} \\
  & 0.541 (0.006) & 0.542 (0.006) \\ \hline
\textbf{Prompt template} & \textbf{No metadata} & \textbf{With metadata} \\
 &  0.541 (0.006) &  0.544 (0.006) \\
\hline
\end{tabular}}
\caption{Ablations on the reasoning dataset using GPT-4.1 as a judge {\includegraphics[height=1em]{llm_clipart.png}}. Average Hellinger distance and standard error across the triplets. Unless stated otherwise, we used the temperature of 0.2 and no metadata as the default prompting condition, which is reported in the main text.}

\resizebox{\columnwidth}{!}{
\begin{tabular}{lcc}
\toprule
\textbf{Temperature} & \textbf{Low (0.2)} & \textbf{High (0.8)} \\
  & 0.374 (0.029) & 0.444 (0.022) \\ \hline
\textbf{Prompt template} & \textbf{No metadata} & \textbf{With metadata} \\
 & 0.374 (0.029) & 0.455 (0.023) \\
\hline
\end{tabular}}\caption{Ablations on the feedback dataset using GPT-4.1 as a judge {\includegraphics[height=1em]{llm_clipart.png}}. Average Hellinger distance and standard error across the triplets. Unless stated otherwise, we used the temperature of 0.2 and no metadata as the default prompting condition, which is reported in the main text.}\label{appendix:prompting-parameters}
\end{table}

\section{Performance of LLM-as-judges methods by task difficulty.}\label{appendix:llm-as-judge-performance} 
\begin{figure}[!h]
  \includegraphics[width=\columnwidth]{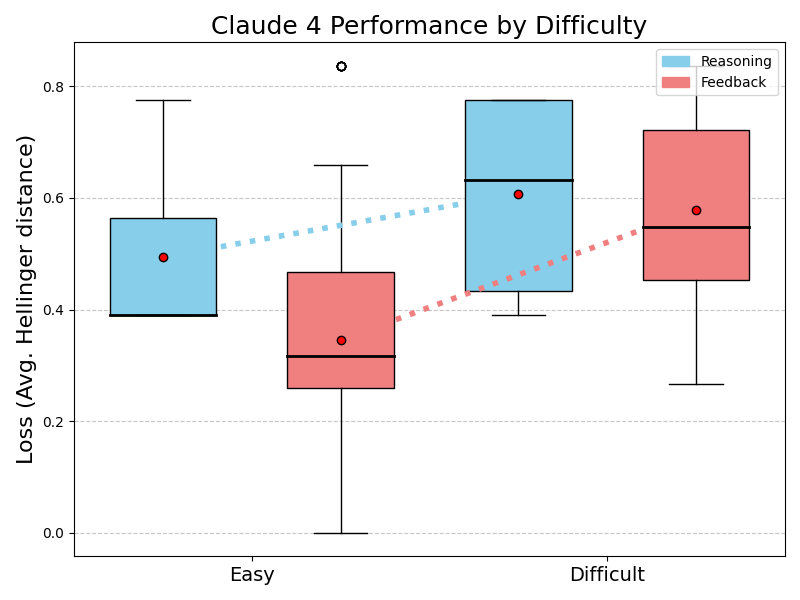}
  \caption{\small \textbf{LLM-as-judges (Claude Sonnet 4) performance variation on easy and difficult task subsets of reasoning and feedback datasets.}}
  \label{fig:model_diff}
\end{figure}

See Fig. \ref{fig:model_diff} and Table \ref{appendix:rq3-performance}.

\begin{table}[h!]
\centering
\resizebox{\columnwidth}{!}{
\begin{tabular}{lcc}
\toprule
\textbf{Method} & \textbf{Reasoning (Hard)} & \textbf{Reasoning (Easy)} \\
\hline
\textbf{LLM-as-judges (GPT-4.1)} {\includegraphics[height=1em]{llm_clipart.png}} & \textbf{0.350 (0.107)} & \textbf{0.734 (0.050)} \\
\textbf{LLM-as-judges (Sonnet 4)} {\includegraphics[height=1em]{llm_clipart.png}} & \textbf{0.350 (0.107)} & \textbf{0.722 (0.050)} \\
\hline
\end{tabular}}

\resizebox{\columnwidth}{!}{
\begin{tabular}{lcc}
\toprule
\textbf{Method} & \textbf{Feedback (Hard)} & \textbf{Feedback (Easy)} \\
\hline
\textbf{LLM-as-judges (GPT-4.1)} {\includegraphics[height=1em]{llm_clipart.png}} & \textbf{0.25 (0.097)} & \textbf{0.696 (0.052)} \\
\textbf{LLM-as-judges (Sonnet 4)} {\includegraphics[height=1em]{llm_clipart.png}} & \textbf{0.25 (0.097)} & \textbf{0.646 (0.054)} \\
\hline
\end{tabular}}
\caption{\small \textbf{Model performance in terms of accuracy on the easy and difficult task subsets.} Higher accuracy means better agreement with majority expert answers. When there are ties in the expert majority answers, we give the model a score of 1 if it matches either of the majority answers, but if the there are ties in the model answers, and there's a single expert majority answer, then we give a score of 0. We report the mean and the standard error. No methods other than LLM-as-a-judge show significant performance difference between the two subgroups split by the expert-perceived difficulty.}\label{appendix:rq3-performance}
\end{table}

\section{Style perturbation experiment.}\label{appendix:style_examples}
\begin{itemize}
\item Original document \#1: Students are debating if 10x10 is 100 or 20., Ultimately they say that it is 100, and that 10x10x10 is 1000. Then they state that 1000 times 100 is `one thousand one hundred`, That $10^{0}$ is 10., Students are counting 10s by 1's, 2's and 3's. The number that they count is the exponent.
\item After style change \#1: Students are debating if 10 multiplied by 10 should be 100 or 20. They conclude that it should be 100. Similarly, they discuss that 10 multiplied by 10, multiplied by another 10 is a thousand. Then, they reach the conclusion that 1000 times 100 is 'one thousand and one hundred.', Students argue that 10 to the power of 0 is 10., Students are counting the powers of 10.
\item Original document \#2: Student debate over 10 raised to 2 being 10 times 2 = 20 and 10 times 10 = 100., That after expanding the expressions you would have to add instead of multiplying.
\item After style change \#2: Students are debating whether 10 to the power of 2 should be interpreted as $10 \times 2 = 20$ or as $10 \times 10$ = 100., Afterwards, students are discussing whether the correct step is to add rather than multiply.
\end{itemize}

\clearpage 
\section{Details about the educational tasks and datasets}\label{appendix:data_collection}
\subsection{Assessing students' mathematical reasoning} 
\paragraph{Annotation process} Eleven expert math teachers were provided with four deidentified transcripts of 6-8th grade math lessons\footnote{These transcripts were collected from middle schools in a large urban public school district in California, under an IRB for human subject research \#75294. The teachers, students, and student guardians consented to their participation.} along with the purpose of the math problems on which students worked.  Each transcript contained between 300-600 lines of student and teacher utterances. Annotators marked specific lines of student utterances from each transcript to answer the following questions: \textit{(1) What are students saying in the selected piece(s) of evidence? (2) What does this piece (or pieces) of evidence tell you about students' understanding and/or progress towards the lesson's purpose?} We also prompted ChatGPT and Claude 3.5 Sonnet to complete the same task and included two outputs per model. The average annotation length is 139.38 words, with a maximum length of 556.

\paragraph{\emph{Odd-one-out} tasks with experts} The dataset of open-ended transcript annotations was then labeled for relative similarity by 10 additional math educators. This resulted in a total of 640 triplet comparisons which we used as benchmarks for evaluating automated measures. The recruited and compensated experts annotated and labeled only deidentified classroom transcript data.

\subsection{Providing feedback to student writing}
\paragraph{Annotation process} We used 18 middle and high school student essays from open-source datasets~\citep{hamner2012asap}, with in-line feedback comments written by 32 experienced English Language Arts (ELA) educators, collected in prior work \cite{mah5213040sentence, tan2024reframing}. We additionally prompted four state-of-the-art models (Chat GPT, Gemini 2.5 Pro, Claude 3.5 Sonnet, and Deepseek-V2.5) to \textit{provide in-line feedback to help students revise their current essay based on the teacher's suggestions}. The average combined length of all feedback comments from a feedback provider was 142.98 words, with a maximum length of 477.

\paragraph{\emph{Odd-one-out} tasks with experts} The dataset of 458 points was labeled by seven ELA teachers. Teachers used the web-based interface, where they were presented with triplets of deidentified feedback either generated by human experts, LLMs, or a mix of both, and instructed to identify the most different one based on the content focus of the substantive changes recommended, rather than the tone, mechanics or style of the comments. 

\subsection{Expert recruitment for annotation and labeling of the de-identified artifacts}We recruited expert ELA and math educators for labeling the similarity data through a graduate school of education alumni mailing list and snowball sampling based on teacher networks affiliated with the school of education. The recruited and compensated experts annotated and labeled only deidentified classroom transcript data. Annotators were compensated at an hourly rate of \$50 for their expertise. Our instructions explained that the collected annotations would be used for research purposes. Demographics of the expert labelers for the student reasoning data are: 4 White or Caucasian female, 2 Asian male, 1 Black or African American male, 1 White or Caucasian male, 1 Multi-racial female, and 1 Multi-racial prefer-not-to-say. Demographics of the expert labelers for the feedback data are: 4 White or Caucasian female, 1 White or Caucasian male, 1 Hispanic or Latine female, and 1 Black or African American female. One ELA educator has 5-6 years of teaching experience and the rest have 8 or more years of experience. They are all based in the U.S.

\subsection{Data accessibility}We share the full annotations for the math reasoning data which we collected under an IRB for human subject research, and a subset of the feedback collected by \citet{mah5213040sentence, tan2024reframing} who consented to sharing a subset of their feedback data to help the reviewers understand the similarity evaluation tasks. The data files are organized as: \texttt{data} which includes the expert and LLM annotations, \texttt{metadata} which includes metadata about the math lesson or the student's writing assignment, \texttt{tasker2task mapping} which contains a mapping of the expert IDs to the task IDs, and \texttt{task2annotation mapping}, which contains a mapping of the task IDs used during the similarity evaluations to the annotation texts. \texttt{similarity judgments} include the labels from the \textit{pick-the-odd-one} evaluations linked to each tasker ID. Aggregated similarity scores are in \texttt{human-eval scores}.

\subsection{Web-based task interfaces}\label{appendix:task_description}
\begin{figure}[htbp]
  \includegraphics[width=\columnwidth]{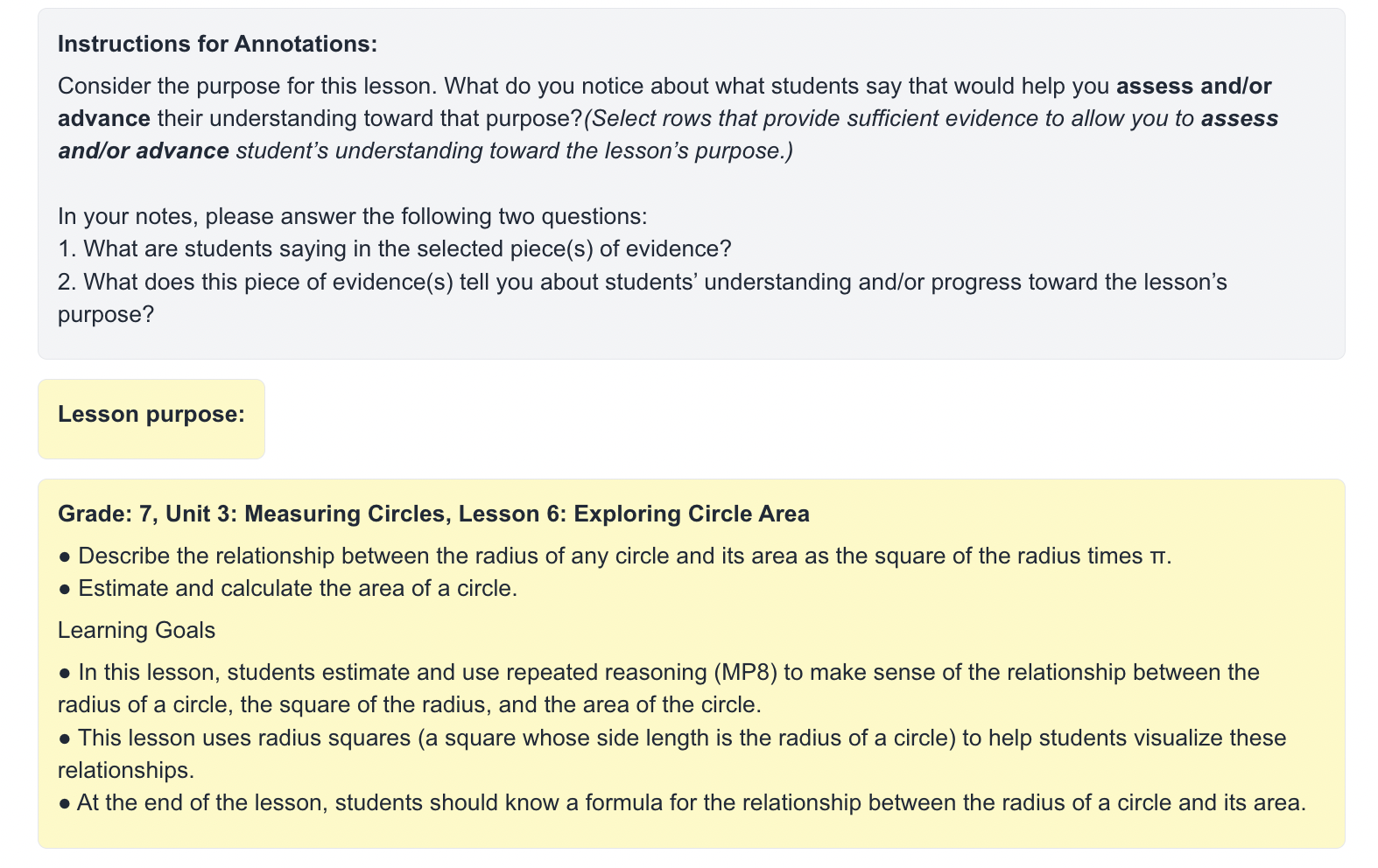}
   \includegraphics[width=\columnwidth]{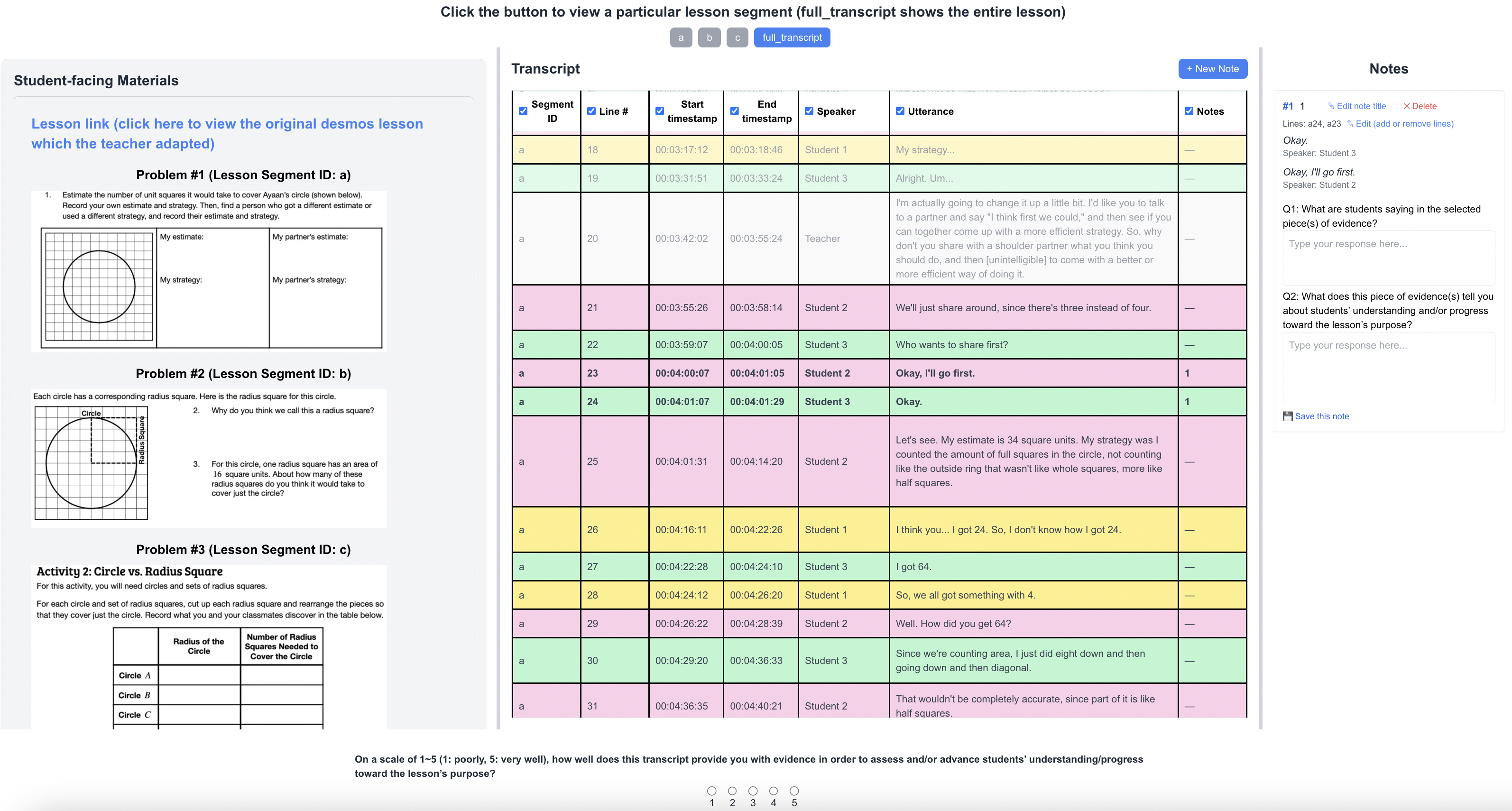}
  \caption{\textbf{Task interface for assessing and interpreting student's mathematical understanding from classroom transcripts.} We collected annotations from expert math educators using a web-based interface. Participants were presented with a classroom transcript, metadata about the lesson's purpose, learning goals, and math activities. Their task was to identify specific lines of student talk that signaled meaningful moments of mathematical reasoning and answer two questions regarding the observed student's understanding and progress towards the learning goals.}
  \label{fig:mol_task}
\end{figure} 

\begin{figure}[htbp]
  \includegraphics[width=\columnwidth]{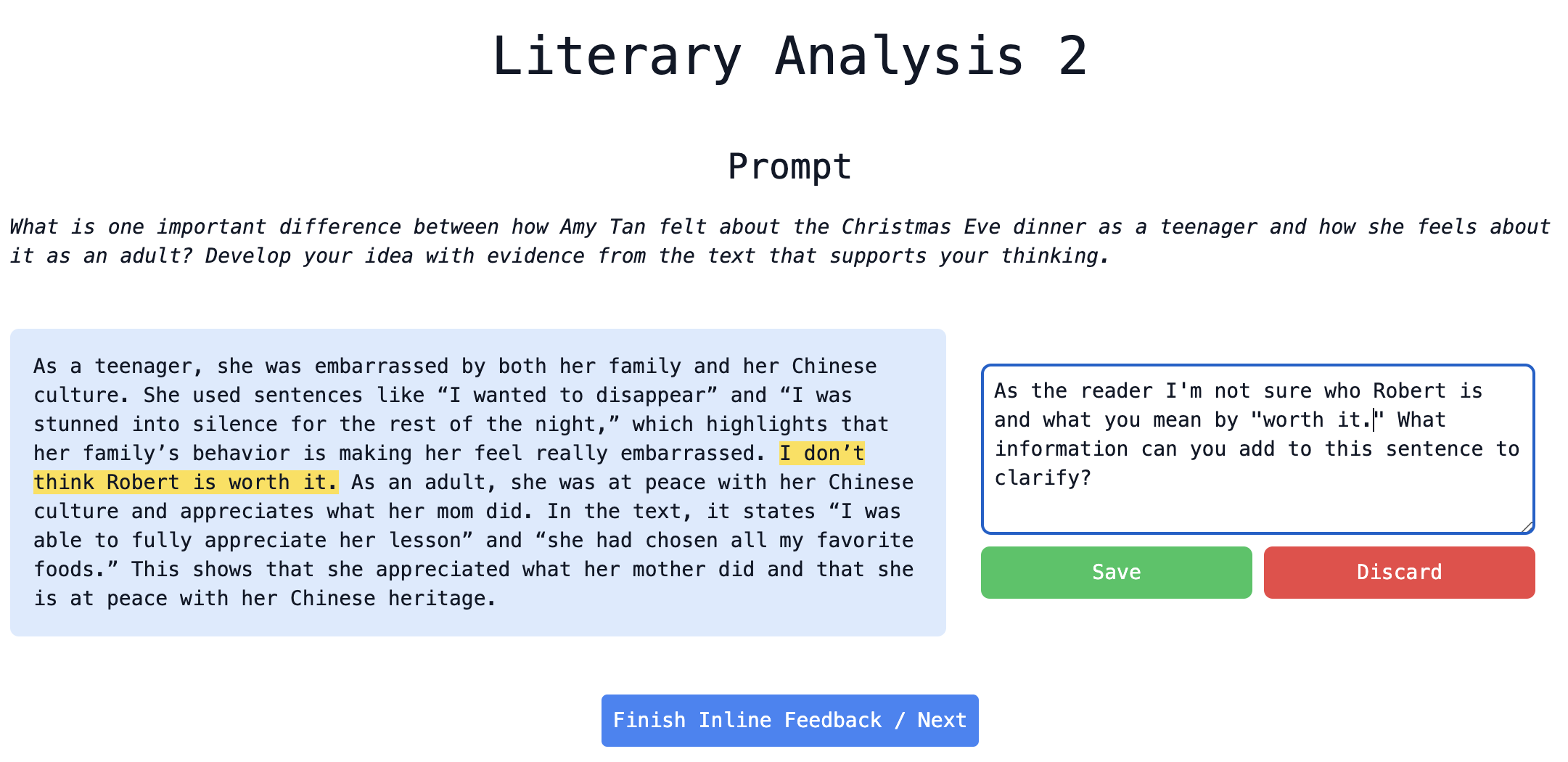}
   \includegraphics[width=\columnwidth]{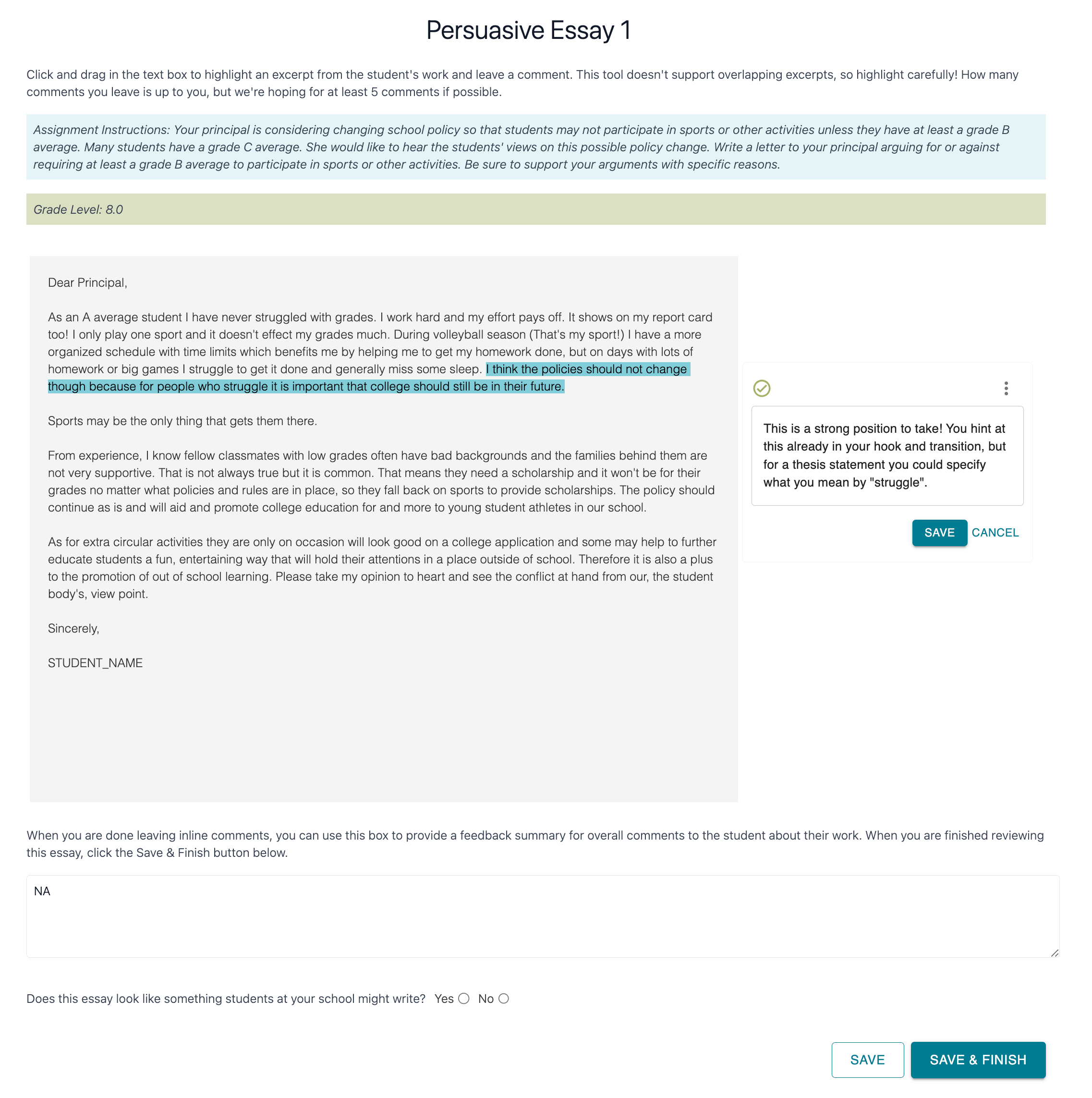}
  \caption{\textbf{Task interface for writing feedback about student's essay by \citet{mah5213040sentence, tan2024reframing}}.}
  \label{fig:feedback_task}
\end{figure} 


\clearpage
\subsection{Expert instruction for the \textit{odd-one-out} task}\label{appendix:prompts}
LLM-as-a-judge system message started with either ``You are a helpful high school English Language Arts teaching assistant" or ``You are a helpful math teacher assistant." The relative judgment task was given in the system prompt and the triplet samples were given in the user message. We added ``Do not include any explanation in your answer" to the system message to enable easier model output processing.

\begin{tcolorbox}[colback=blue!10, colframe=black, coltitle=black, colbacktitle=white, title=Math expert evaluation instruction, fonttitle=\bfseries]
Teachers were asked to make notes about what students said in the transcript that would help them assess and/or advance students' understanding towards the lesson purpose. These are the notes that three different teachers made about ``what students are saying'' in the pieces of transcript. Select which note is most different in terms of the underlying ideas (rather than focusing on the style, tone, or length of the notes).
\end{tcolorbox}

\begin{tcolorbox}[colback=blue!10, colframe=black, coltitle=black, colbacktitle=white, title=ELA expert evaluation instruction, fonttitle=\bfseries]
\{Essay instruction / writing prompt given to the student.\}
Three teachers asked to give the student feedback for revising their writing. Select which feedback is most different in terms of what it asks the student to change in their writing. Focus on the feedback content, rather than judging based on the delivery (e.g., teacher's writing style, tone, length, use of questions versus directives in the feedback).
\end{tcolorbox}

\section{Hyperparameters for topic distribution modeling.}
We used \href{https://maartengr.github.io/BERTopic/getting_started/quickstart/quickstart.html}{BERTopic package} for topic modeling, \href{https://umap-learn.readthedocs.io/en/latest/}{UMAP} for dimension reduction, and \href{https://pypi.org/project/hdbscan/}{HDBSCAN} for clustering. The parameters for UMAP are: n neighbors = 3, n components = 100, min distance = 0, metric = cosine. The parameters for HDBSCAN are min samples = 1, metric = euclidean. We set the min cluster size of HDBSCAN as a tunable hyperparameter and enumerated between $\{2, 3, .., 15\}$ and reported the best result from hyperparameter search. 

We used an open-sourced Python wrapper for MALLET implementation of LDA by \href{https://github.com/maria-antoniak/little-mallet-wrapper}{maria-antoniak/little-mallet-wrapper}. For both the math and the feedback domains, we enumerated the number of topics to find from $\{50, 75, 100, 125, 150\}$ and reported the best result. With the math data, LDA topic modeling could not find more than 125 unique topics.

Our experiments do not rely on GPU resources.
\end{document}